\documentclass[runningheads]{llncs}
\usepackage[T1]{fontenc}
\usepackage{graphicx}
\usepackage{float}
\usepackage{placeins}
\usepackage{color}
\usepackage{amsmath}
\usepackage{fca}

\newcommand{\R}{\mathbb{R}}
\newcommand{\K}{\mathbb{K}}
\newcommand{\Kout}{\mathbb{K}_{\mathrm{outer}}}
\newcommand{\Kin}{\mathbb{K}_{\mathrm{inner}}}
\newcommand{\B}{\mathfrak{B}}
\newcommand{\uB}{\underline{\mathfrak{B}}}
\newcommand{\Mout}{M_{\mathrm{outer}}}
\newcommand{\Min}{M_{\mathrm{inner}}}
\newcommand{\Iout}{I_{\mathrm{outer}}}
\newcommand{\Iin}{I_{\mathrm{inner}}}
\newcommand{\gammaout}{\gamma_{\mathrm{outer}}}
\newcommand{\gammain}{\gamma_{\mathrm{inner}}}
\newcommand{\botout}{\bot_{\mathrm{outer}}}
\newcommand{\botin}{\bot_{\mathrm{inner}}}

\begin{document}

\title{Exploring ESC Winners with Nested Diagrams}

\author{
    Anurag Sharma\inst{1,2} \and
    Marcel~Nöhre\inst{1,2} \and
    Gerd Stumme\inst{1,2}
}

\authorrunning{A. Sharma et al.}

\institute{Knowledge \& Data Engineering Group (KDE), University of Kassel, Germany \and Interdisciplinary Research Center for Information Systems Design (ITeG), University of Kassel, Germany \\
\email{\{sharma,noehre,stumme\}@cs.uni-kassel.de}}

\maketitle

\begin{abstract}
We present \emph{ConceptFlow}, a scikit-learn-compatible Python library for Formal Concept Analysis that constructs and renders nested line diagrams from many-valued formal contexts. Given a many-valued context and a partition of its attributes into conceptual scales, ConceptFlow performs conceptual scaling, computes the factor lattices, identifies filled nodes of the corresponding subdirect product, and produces an interactive visualization.

We apply ConceptFlow to the winners of the Eurovision Song Contest from 1975 to 2025, exploring relationships between voting patterns and musical characteristics. Voting support is captured by an outer scale spanning regional, cultural, historical, and political dimensions, while an inner scale captures musical characteristics via tempo and key. The resulting nested line diagram reveals implications across both scales, exposing dependencies between how winning entries were voted for and the musical properties they share.
\end{abstract}

%%%%%%%%%%%%%%%%%%%%%%%%%%%%%%%%%%%%%%%%%%%%%%%%%%%%%%%%%%%%%%%%%%%%%%%%%%%%%%%%
\section{Introduction}
\label{sec:introduction}
%%%%%%%%%%%%%%%%%%%%%%%%%%%%%%%%%%%%%%%%%%%%%%%%%%%%%%%%%%%%%%%%%%%%%%%%%%%%%%%%
The Eurovision Song Contest (ESC) is a yearly music competition where each participating country enters a song and judges the songs of all other countries. Since 1975, every country distributes the scores 1-8, 10, and 12 to its ten favourite performances, never voting for itself. How those scores are decided, by juries, public televoting, or a combination, has changed repeatedly over the decades. The contest also has a long-standing reputation for biased voting, with persistent alliances between countries. Greece and Cyprus, for instance, reliably award each other 12 points~\cite{Gatherer2006}.

Formal Concept Analysis (FCA) provides a natural framework for studying such relationships, representing voting patterns and musical characteristics within a single conceptual structure. In this paper, we present the nested line diagram functionality of \emph{ConceptFlow}~\cite{ConceptFlow}, a scikit-learn-compatible Python library for Formal Concept Analysis whose components follow the scikit-learn estimator interface. We demonstrate its use by analyzing ESC winners. The outer scale captures voting support across taxonomic clusters, while the inner scale captures musical characteristics. The resulting formal context reveals implications between these two dimensions, exposing dependencies between how winning entries were voted for and the musical properties they share.
%%%%%%%%%%%%%%%%%%%%%%%%%%%%%%%%%%%%%%%%%%%%%%%%%%%%%%%%%%%%%%%%%%%%%%%%%%%%%%%%
\section{Background}
\label{sec:background}
%%%%%%%%%%%%%%%%%%%%%%%%%%%%%%%%%%%%%%%%%%%%%%%%%%%%%%%%%%%%%%%%%%%%%%%%%%%%%%%%
We assume familiarity with FCA~\cite{GanterFCA2024}. Briefly, a \emph{formal context} $\GMI$ consists of a set $G$ of objects, a set $M$ of attributes, and a relation $I \subseteq G \times M$. A \emph{formal concept} is a pair $(A, B)$ of a maximal object set and the attributes they share; ordered by extent inclusion, these concepts form the \emph{concept lattice} $\BVGMI$.

Let $\K = \Kout \mid \Kin$ denote the apposition of two formal contexts. The concept lattice $\uB(\K)$ is isomorphic to a subdirect join-semilattice product of the factor lattices $\uB(\Kout)$ and $\uB(\Kin)$~\cite{GanterFCA2024}. Thus, every formal concept $(A, B)~\in~\B(\K)$ is represented by exactly one pair consisting of an outer concept and an inner concept.

A \emph{line diagram} of $\BVGMI$ represents each formal concept by a node and connects two nodes by a straight segment when $c_1 \leq c_2$ is in the transitive reduction of $\leq$, drawing the smaller concept strictly below the larger. Each node is assigned coordinates in $\R^2$, which determine the position of the corresponding concept in the drawing. \emph{Doubly-additive} line diagrams form a special class: each concept's position in $\R^2$ is the sum of the vectors of the objects in its extent and the attributes in its complement intent, where positive $y$-components ensure the vertical order respects the order relation $\leq$. It suffices to assign vectors to the irreducibles of the reduced context. In this work, we consider \textbf{DimFlux}~\cite{Noehre2026}, which projects \emph{DimDraw} coordinates into the additive space and refines them for readability using a force-based model.

\emph{Nested line diagrams} visualize the concept lattice of a formal context by decomposing its attribute set into several \emph{scales}. Rather than constructing a single concept lattice over all attributes, the context is viewed as the apposition of subcontexts sharing the same object set. One factor lattice is displayed as the \emph{outer} diagram, while a copy of another factor lattice is placed inside each of its concepts~\cite{GanterFCA2024}. The construction of such a nested line diagram is described in detail in Section~\ref{sec:nested-construction}.
%%%%%%%%%%%%%%%%%%%%%%%%%%%%%%%%%%%%%%%%%%%%%%%%%%%%%%%%%%%%%%%%%%%%%%%%%%%%%%%%
\section{ESC Dataset and Scale Construction}
\label{sec:dataset-scales}
%%%%%%%%%%%%%%%%%%%%%%%%%%%%%%%%%%%%%%%%%%%%%%%%%%%%%%%%%%%%%%%%%%%%%%%%%%%%%%%%
We analyse the 50 winners of the Eurovision Song Contest from 1975 to 2025, excluding the cancelled 2020 contest. Voting results and song metadata come from the publicly available Eurovision Song Contest Dataset\footnote{\url{https://github.com/EurovisionAPI/dataset}}. For each winner, we collect the final voting results\footnote{Before 2016, the ESC reports only a combined voting result. Since 2016, jury and public televoting are reported separately. For consistency, we represent each winner by a single voting profile: the combined score before 2016, and thereafter whichever of jury or televote awards the larger total number of points.} together with musical metadata, including the song's BPM and key. The resulting many-valued context forms the input for the conceptual scaling process described below.

The outer scale captures whether a winning entry received strong support from countries to which it has strong ties. We studied four different types of ties: \emph{regional}, \emph{cultural}, \emph{historical}, and \emph{political}. For each of the four types, we manually clustered the set of participating countries, as described in our blog\footnote{\url{https://www.kde.cs.uni-kassel.de/blogs/esc}}, to reflect commonly discussed sources of voting affinity~\cite{Gatherer2006,Yair1995}. For each winning entry $g$ and cluster $m$ in any of the four clusterings, we set $(g, m) \in I$ if winner $g$ received at least eight points on average from the eligible countries in $m$.

The inner scale captures two musical properties of the winning songs: tempo and key. Tempo is transformed using a threshold scale with thresholds at 100 and 150 BPM, yielding the Boolean attributes \emph{$\geq$~100 BPM} and \emph{$\geq$~150 BPM}~\cite{Madison2010}. Since the scale is ordinal, every song with tempo at least 150 BPM also satisfies \emph{$\geq$~100 BPM}. Key is represented by a dichotomic scale with the mutually exclusive attributes \emph{major} and \emph{minor}.
%%%%%%%%%%%%%%%%%%%%%%%%%%%%%%%%%%%%%%%%%%%%%%%%%%%%%%%%%%%%%%%%%%%%%%%%%%%%%%%%
\section{Constructing the Nested Line Diagram}
\label{sec:nested-construction}
%%%%%%%%%%%%%%%%%%%%%%%%%%%%%%%%%%%%%%%%%%%%%%%%%%%%%%%%%%%%%%%%%%%%%%%%%%%%%%%%
In this work, we present \emph{ConceptFlow}, a scikit-learn-compatible Python library for Formal Concept Analysis providing FCA basics, conceptual scaling, and visualization tools~\cite{ConceptFlow}. Given a many-valued formal context (such as the ESC dataset described in Section~\ref{sec:dataset-scales}), we independently construct the outer and inner formal contexts by conceptual scaling and compute their corresponding concept lattices.

%%%%%%%%%%%%%%%%%%%%%%%%%%%%%%%%%%%%%%%%%%%%%%%%%%%%%%%%%%%%%%%%%%%%%%%%%%%%%%%%
\subsection{Building the Factor Lattices}
\label{subsec:factor-lattices}
%%%%%%%%%%%%%%%%%%%%%%%%%%%%%%%%%%%%%%%%%%%%%%%%%%%%%%%%%%%%%%%%%%%%%%%%%%%%%%%%
The construction begins with a many-valued context whose object set $G$ consists of the 50 ESC winners. Conceptual scaling is then applied to derive two formal contexts, $\Kout = (G, \Mout, \Iout)$ and $\Kin = (G, \Min, \Iin)$, where $\Mout$ contains all country clusters of the four clusterings, while \linebreak $\Min = \{\geq 100\,\textrm{BPM}, \geq 150\,\textrm{BPM}, \textrm{minor}, \textrm{major}\}$. 

The concept lattices of $\Kout$ and $\Kin$ provide the factor lattices of the nested line diagram. Since both contexts share the same object set, they satisfy the apposition condition described in Section~\ref{sec:background}.

%%%%%%%%%%%%%%%%%%%%%%%%%%%%%%%%%%%%%%%%%%%%%%%%%%%%%%%%%%%%%%%%%%%%%%%%%%%%%%%%
\subsection{Computing Filled Nodes}
\label{subsec:filled-nodes}
%%%%%%%%%%%%%%%%%%%%%%%%%%%%%%%%%%%%%%%%%%%%%%%%%%%%%%%%%%%%%%%%%%%%%%%%%%%%%%%%
Rather than constructing the concept lattice of the full apposition explicitly, we compute the filled nodes directly from the two factor lattices. For every object $g \in G$, we determine its object concepts $\gammaout(g)$ and $\gammain(g)$, which form the initial set of atomic coordinate pairs
\[
P_0 = \left\{ (\gammaout(g), \gammain(g)) \mid g \in G \right\} \cup \left\{ (\botout, \botin) \right\}.
\]
We add $(\botout, \botin)$ explicitly, since it represents $(M',M)$, which is always included in the subdirect product. Equivalently, it is the join (supremum) of the empty set and therefore cannot be generated from the atomic pairs by taking non-empty joins.

We then repeatedly compute the componentwise joins of all coordinate pairs until no new pairs are generated. The resulting fixpoint is exactly the image of the subdirect join-preserving embedding
\[
\varphi : \uB(\K) \rightarrow \uB(\Kout) \times \uB(\Kin),
\]
identifying the inner formal concepts to be filled in the nested line diagram.

%%%%%%%%%%%%%%%%%%%%%%%%%%%%%%%%%%%%%%%%%%%%%%%%%%%%%%%%%%%%%%%%%%%%%%%%%%%%%%%%
\subsection{Rendering}
\label{subsec:rendering}
%%%%%%%%%%%%%%%%%%%%%%%%%%%%%%%%%%%%%%%%%%%%%%%%%%%%%%%%%%%%%%%%%%%%%%%%%%%%%%%%
After the filled nodes have been determined, both factor lattices are laid out using DimFlux~\cite{Noehre2026}. The outer lattice is drawn once, while the inner lattice's layout is computed only once and then reused as a fixed template inside every outer concept. As a result, corresponding inner concepts occupy identical relative positions throughout the visualization, so that differences between outer concepts are visible solely through which nodes are filled.

The complete nested line diagram is then exported as a JSON representation containing node positions, edges, labels, and filled coordinate pairs. This representation is rendered as an interactive D3.js~\cite{Bostock2011} visualization, cleanly separating the visualization layer from the underlying FCA implementation, so that the rendering component remains independent of the construction algorithms.

%%%%%%%%%%%%%%%%%%%%%%%%%%%%%%%%%%%%%%%%%%%%%%%%%%%%%%%%%%%%%%%%%%%%%%%%%%%%%%%%
\section{Results}
\label{sec:results}
%%%%%%%%%%%%%%%%%%%%%%%%%%%%%%%%%%%%%%%%%%%%%%%%%%%%%%%%%%%%%%%%%%%%%%%%%%%%%%%%
\begin{figure}[t]
    \centering
    \includegraphics[width=1\textwidth]{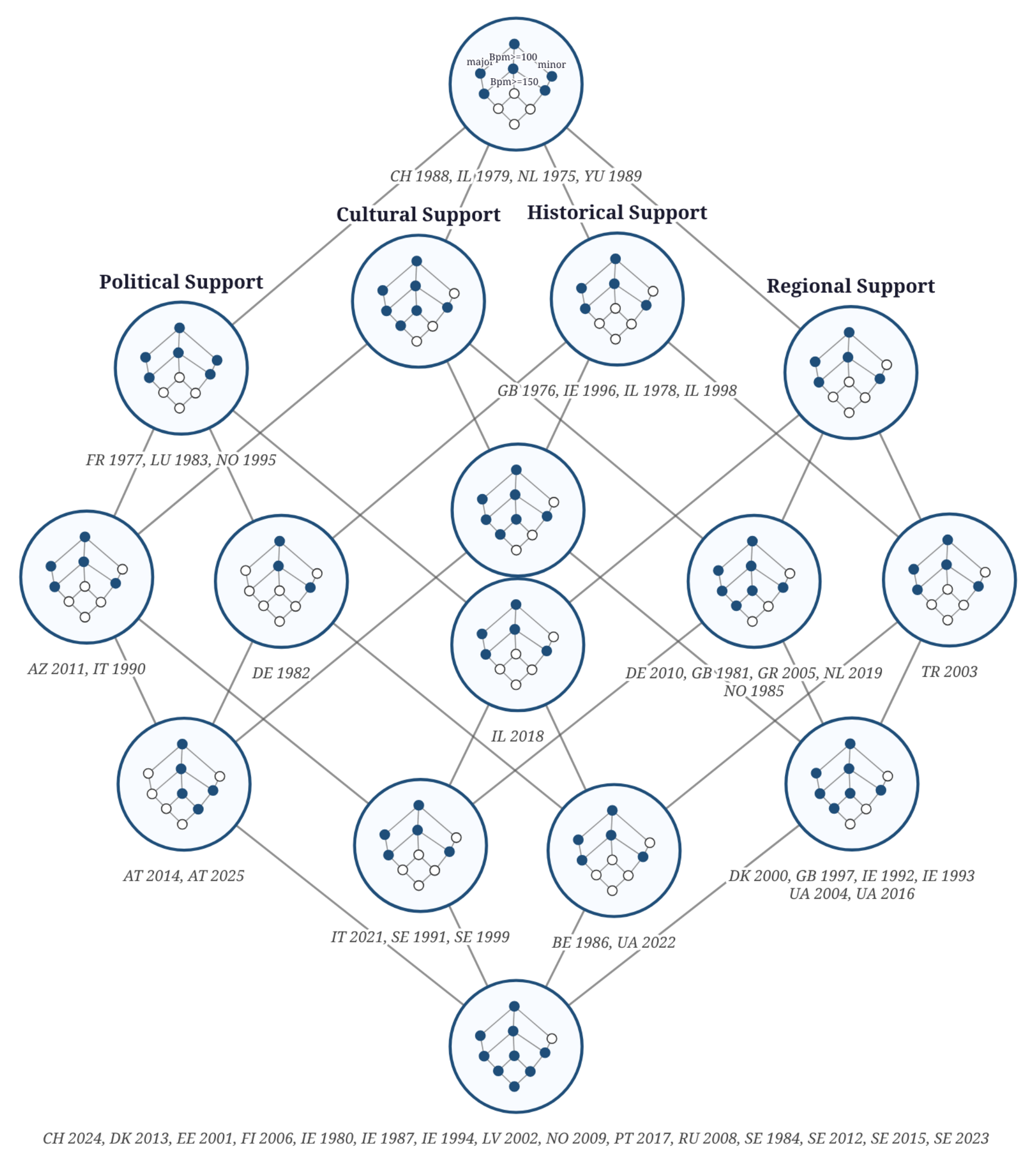}
    \caption{Nested line diagram of the ESC winners generated with ConceptFlow. Filled nodes represent concepts of the original context, whereas hollow nodes correspond to implications visible in the combined context.}
    \label{fig:outer-lattice}
\end{figure}

Filled nodes represent the concepts of the original context after decomposing it into outer and inner scales. Unfilled nodes correspond to implications. To determine what an unfilled node implies, one locates the unique largest filled node below it, either within the same inner lattice or, if necessary, in a lower outer concept. Since the attributes of the unfilled node are contained in those of this filled node, the additional attributes introduced constitute the implication.

The nested line diagram (Figure~\ref{fig:outer-lattice})\footnote{An interactive version of the nested line diagram and a companion blog post with a more detailed discussion of the methodology, results, and implications are available at \url{https://www.kde.cs.uni-kassel.de/blogs/esc}.} makes several implications between voting patterns and musical characteristics visible that hold in the analyzed dataset. Among the 50 ESC winners considered, every winning song with a tempo of at least $150$ BPM also received cultural support. Another dependency relates to tempo and key mode. Among winners receiving regional or cultural support, every minor-key song has a tempo of at least $100$ BPM. This is reflected by the absence of the corresponding node for a slow minor-key song, whereas the node representing minor-key songs with a tempo of at least $100$ BPM is filled.

The diagram also reveals relationships involving combinations of voting-support attributes. In particular, winners that combine cultural support with a minor key and a tempo of at least $150$ BPM also receive political and historical support. The absence of the corresponding node in the nested diagram therefore indicates that this combination never occurs without the additional voting-support attributes.

These observations illustrate how the nested representation localises implications within particular nodes of the subdirect product. Rather than listing implications algebraically, the diagram shows where musical combinations are possible and where they are excluded by additional voting-support conditions.

%%%%%%%%%%%%%%%%%%%%%%%%%%%%%%%%%%%%%%%%%%%%%%%%%%%%%%%%%%%%%%%%%%%%%%%%%%%%%%%%
\section{Conclusion}
\label{sec:conclusion}
%%%%%%%%%%%%%%%%%%%%%%%%%%%%%%%%%%%%%%%%%%%%%%%%%%%%%%%%%%%%%%%%%%%%%%%%%%%%%%%%
We presented \emph{ConceptFlow}, a scikit-learn-compatible Python library for Formal Concept Analysis supporting conceptual scaling, concept lattice construction, and interactive nested line diagrams. Using the Eurovision Song Contest as a case study, we demonstrated how nested line diagrams combine multiple conceptual scales into a single visualization, making relationships and implications between voting patterns and musical characteristics directly accessible. The ESC analysis serves as a demonstration of the library's capabilities rather than as a comprehensive statistical study of ESC voting behaviour.

Future work will extend ConceptFlow with additional conceptual scales, support for implication theory, and further interactive visualization capabilities. In particular, we plan to compute canonical implication bases and integrate them into the interactive interface, allowing users to inspect the implications associated with concepts and unfilled regions of the diagrams directly within the visualization. This will further strengthen the connection between visualization and knowledge discovery in Formal Concept Analysis.

\bibliographystyle{splncs04}
\bibliography{bibliography}

@article{Gatherer2006,
   title = {Comparison of Eurovision Song Contest Simulation with Actual Results Reveals Shifting Patterns of Collusive Voting Alliances.},
   author = {Gatherer, Derek},
   journal = {Journal of Artificial Societies and Social Simulation},
   ISSN = {1460-7425},
   volume = {9},
   number = {2},
   pages = {1},
   year = {2006},
   URL = {https://www.jasss.org/9/2/1.html},
}

@book{GanterFCA2024,
	title         = {Formal Concept Analysis - Mathematical Foundations},
	subtitle      = {Mathematical Foundations},
	author        = {Bernhard Ganter and Rudolf Wille},
	publisher     = {Springer},
	year          = {2024},
	edition       = {2nd},
	pages         = {XII, 370},
	isbn          = {978-3-031-63421-5},
	ebookisbn     = {978-3-031-63422-2},
	softcoverisbn = {978-3-031-63424-6},
}

@article{Noehre2026,
  author  = {Marcel Nöhre and Dominik Dürrschnabel and Bernhard Ganter and Gerd Stumme},
  title   = {{DimFlux}: Force-Directed Additive Line Diagrams},
  journal = {International Journal of Approximate Reasoning},
  volume  = {197},
  pages   = {109734},
  year    = {2026},
  issn    = {0888-613X},
  doi     = {10.1016/j.ijar.2026.109734},
  url     = {https://www.sciencedirect.com/science/article/pii/S0888613X2600109X}
}

@article{Bostock2011,
  author  = {Mike Bostock and Vadim Ogievetsky and Jeffrey Heer},
  title   = {D3: Data-Driven Documents},
  journal = {IEEE Transactions on Visualization and Computer Graphics},
  volume  = {17},
  number  = {12},
  pages   = {2301--2309},
  year    = {2011},
  doi     = {10.1109/TVCG.2011.185}
}

@article{Yair1995,
  author  = {Gideon Yair},
  title   = {{`Unite Unite Europe'}: The Political and Cultural Structures of Europe as Reflected in the Eurovision Song Contest},
  journal = {Social Networks},
  volume  = {17},
  number  = {2},
  pages   = {147--161},
  year    = {1995},
  doi     = {10.1016/0378-8733(95)00253-K},
  url     = {https://www.sciencedirect.com/science/article/pii/037887339500253K}
}

@article{Madison2010,
  author  = {Guy Madison and Johan Paulin},
  title   = {Ratings of Speed in Real Music as a Function of Both Original and Manipulated Beat Tempo},
  journal = {The Journal of the Acoustical Society of America},
  volume  = {128},
  number  = {5},
  pages   = {3032--3040},
  year    = {2010},
  month   = nov,
  doi     = {10.1121/1.3493462},
  pmid    = {21110598},
  url      = {https://pubmed.ncbi.nlm.nih.gov/21110598/}
}

@misc{ConceptFlow,
  author       = {Anurag Sharma},
  title        = {{ConceptFlow}: A Scikit-Learn-Compatible {Python} Library for Formal Concept Analysis},
  year         = {2026},
  howpublished = {\url{https://github.com/anuragxorma/conceptflow}},
  note         = {{GitHub} repository}
}

\end{document}